\documentclass[10pt,twocolumn,letterpaper]{article}

\usepackage{iccv}
\usepackage{times}
\usepackage{epsfig}
\usepackage{graphicx}
\usepackage{amsmath}
\usepackage{amssymb}
\usepackage{booktabs}
\usepackage{threeparttable}
\usepackage{multirow}
\usepackage{lipsum} 
\usepackage[table,xcdraw]{xcolor}
\usepackage{pdfpages}

\usepackage[pagebackref=true,breaklinks=true, colorlinks,bookmarks=false]{hyperref}

\iccvfinalcopy 

\def\iccvPaperID{2126} 

\ificcvfinal\fi
\begin{document}

\title{Semi-Supervised Video Salient Object Detection Using Pseudo-Labels}

\author{
    Pengxiang Yan$^{1}$ \quad
    Guanbin Li$^{1}$\thanks{Corresponding author is Guanbin Li.} \quad
    Yuan Xie$^{1,2}$ \quad\\
    Zhen Li$^{3}$  \quad
    Chuan Wang$^{4}$ \quad
    Tianshui Chen$^{1,2}$ \quad
    Liang Lin$^{1,2}$\vspace{1mm}\quad\\
    $^1$Sun Yat-sen University \quad
    $^2$DarkMatter AI Research \quad
    $^4$Megvii Technology \quad\\
    $^3$Shenzhen Research Institute of Big Data, the Chinese University of Hong Kong (Shenzhen) \\
    {\tt\small yanpx@mail2.sysu.edu.cn},
    {\tt\small liguanbin@mail.sysu.edu.cn},
    {\tt\small xiey39@mail2.sysu.edu.cn},\\
    {\tt\small lizhen36@connect.hku.hk},
    {\tt\small wangchuan@megvii.com},
    {\tt\small tianshuichen@gmail.com},
    {\tt\small linliang@ieee.org}
}

\maketitle
\thispagestyle{empty}

\begin{abstract}
    Deep learning-based video salient object detection has recently achieved great success with its performance significantly outperforming any other unsupervised methods. However, existing data-driven approaches heavily rely on a large quantity of pixel-wise annotated video frames to deliver such promising results. In this paper, we address the semi-supervised video salient object detection task using pseudo-labels. Specifically, we present an effective video saliency detector that consists of a spatial refinement network and a spatiotemporal module. Based on the same refinement network and motion information in terms of optical flow, we further propose a novel method for generating pixel-level pseudo-labels from sparsely annotated frames. By utilizing the generated pseudo-labels together with a part of manual annotations, our video saliency detector learns spatial and temporal cues for both contrast inference and coherence enhancement, thus producing accurate saliency maps. Experimental results demonstrate that our proposed semi-supervised method even greatly outperforms all the state-of-the-art fully supervised methods across three public benchmarks of VOS, DAVIS, and FBMS.
\end{abstract}

\section{Introduction}
Salient object detection aims at identifying the most visually distinctive objects in an image or video that attract human attention.
In contrast to the other type of saliency detection, i.e., eye fixation prediction~\cite{kruthiventi2017deepfix, wang2018revisiting} which is designed to locate the focus of human attention, salient object detection focuses on segmenting the most salient objects with precise contours.
This topic has drawn widespread interest as it can be applied to a wide range of vision applications, such as object segmentation~\cite{wei2017stc}, visual tracking~\cite{wu2014weighted}, video compression~\cite{itti2004automatic}, and video summarization~\cite{ma2002user}.

Recently, video salient object detection has achieved significant progress~\cite{li2018flow, song2018pyramid, wang2018video} due to the development of deep convolutional neural networks~(CNNs). However, the performance of these deep learning-based methods comes at the cost of a large quantity of densely annotated frames. It is arduous and time consuming to manually annotate a large number of pixel-level video frames since even an experienced annotator needs several minutes to label a single frame. Moreover, a video clip usually contains hundreds of video frames with similar content. To reduce the impact of label noise on model training, the annotators need to spend considerable time checking the consistency of the label before and after. Considering that visual saliency is subjective, the annotation work becomes even more difficult, and the quality of the labeling is hard to guarantee.


\begin{figure}[t]
    \begin{center}
       \includegraphics[width=0.997\linewidth]{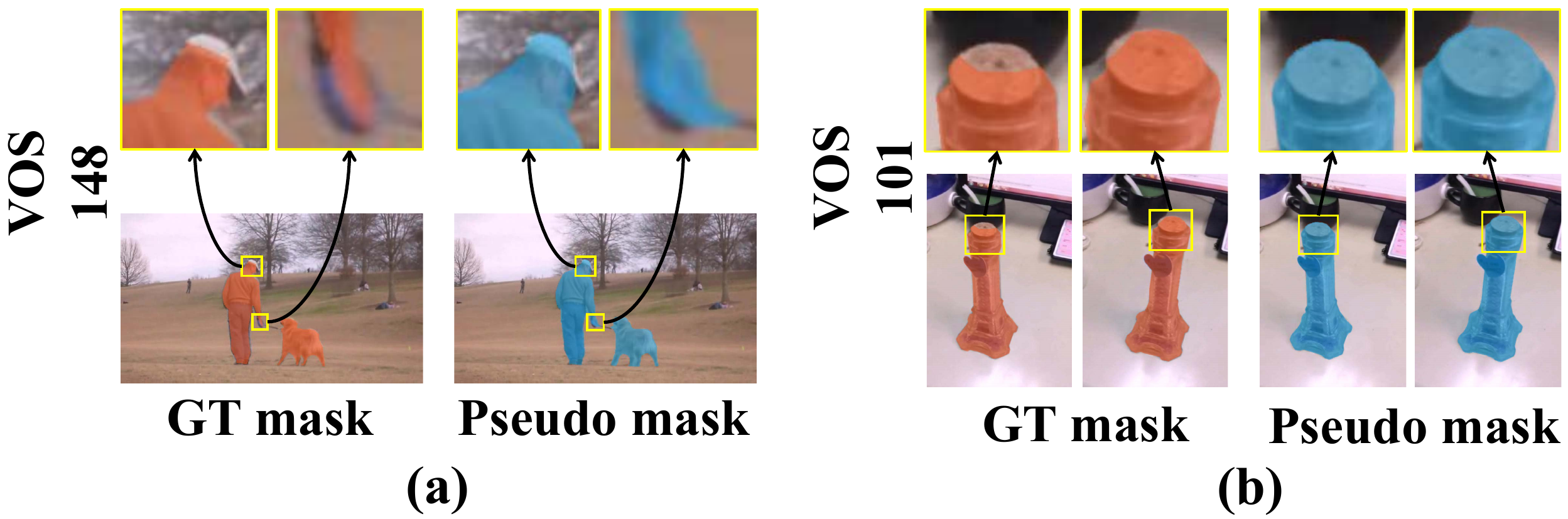}
    \end{center}
    \vspace{-4mm}
    \caption{Example ground truth masks (orange mask) vs. our generated pseudo-labels (blue mask) from the VOS~\cite{li2018benchmark} dataset.}
    \vspace{-5mm}
 \label{fig:visual_pseudo_label}
 \end{figure}

Although there are many unsupervised video salient object detection methods~\cite{wang2015saliency, wang2015consistent, li2018benchmark} that are free of numerous training samples, these methods suffer from low prediction accuracy and efficiency. Since most of these methods exploit hand-crafted low-level features, e.g., color, gradient or contrast, they work well in some considered cases while failing in other challenging cases.
Recent research by Li \textit{et al.}~\cite{li2018weakly} noticed the weakness of unsupervised methods and the lack of annotations for deep learning-based methods. They attempted to use the combination of coarse activation maps and saliency maps, which were generated by learning-based classification networks and unsupervised methods respectively, as pixel-wise training annotations for image salient object detection. However, this method is not suitable for the video-based salient object detection task, where object motion and changes in appearance contrast are more attractive to human attention~\cite{itti1998model} than object categories. Moreover, it is also challenging to train deep learning-based video salient object detection models for temporally consistent saliency map generation, due to the lack of temporal cues in sparsely annotated frames.

By carefully observing the training samples of existing video salient object detection benchmarks~\cite{li2018benchmark, perazzi2016benchmark, brox2010object}, we found that the adjacent frames in a video share small differences due to the high video sampling rate~(e.g., 24 fps in the DAVIS~\cite{perazzi2016benchmark} dataset). Thus, we conjecture that it is not necessary to densely annotate all the frames since some of the annotations can be estimated by exploiting motion information. Moreover, recent work has shown that a well-trained CNN can also correct some manual annotation errors that exist in the training samples~\cite{li2018weakly}.

Inspired by these observations, in this paper, we address the semi-supervised video salient object detection task using unannotated frames with pseudo-labels as well as a few sparsely annotated frames. We develop a framework that exploits pixel-wise pseudo-labels generated from a few ground truth labels to train a video-based convolutional network for saliency maps with spatiotemporal coherence. Specifically, we first propose a refinement network with residual connections~(RCRNet) to extract spatial saliency information and generate saliency maps with high-resolution through a series of upsampling based refine operations. Then, the RCRNet equipped with a non-locally enhanced recurrent~(NER) module is proposed to enhance the spatiotemporal coherence of the resulting saliency maps. For the pseudo-label generation, we adopt a pretrained FlowNet 2.0~\cite{ilg2017flownet} for motion estimation between labeled and unlabeled frames and propagate adjacent labels to unlabeled frames. Meanwhile, another RCRNet is modified to accept multiple channels as input, including RGB channels, propagated adjacent ground truth annotations, and motion estimations, to generate consecutive pixel-wise pseudo-labels, which make up for the temporal information deficiency that exists in sparse annotations.
As shown in Fig.~\ref{fig:visual_pseudo_label}, our model can produce reasonable and consistent pseudo-labels, which can even improve the boundary details (Example a) and overcome the labeling ambiguity between frames (Example b).
Learning under the supervision of generated pseudo-labels together with a few ground truth labels, our proposed RCRNet with NER module~(RCRNet+NER) can generate more accurate saliency maps which even outperforms the results of top-performing fully supervised video salient object detection methods.

In summary, this paper has the following contributions:
{\flushleft $\bullet$} We introduce a refinement network equipped with a non-locally enhanced recurrent module to generate saliency maps with spatiotemporal coherence. 
{\flushleft $\bullet$} We further propose a flow-guided pseudo-label generator, which captures the interframe continuity of video and generates pseudo-labels of intervals based on sparse annotations.
{\flushleft $\bullet$} Under the joint supervision of the generated pseudo-labels and the manually labeled sparse annotations~(e.g., 20\% ground truth labels), our semi-supervised model can be trained to outperform existing state-of-the-art fully supervised video salient object detection methods.

\section{Related Work}
\subsection{Salient Object Detection}

Benefiting from the development of deep convolutional networks, salient object detection has recently achieved significant progress. In particular, these methods based on the fully convolutional network~(FCN) and its variants~\cite{li2017instance, hou2017deeply, li2016deep} have become the dominant methods in this field, due to their powerful end-to-end feature learning nature and high computational efficiency. Nevertheless, these methods are inapplicable to video salient object detection without considering spatiotemporal information and contrast information within both motion and appearance in videos. Recently, attempts to apply deep CNNs to video salient object detection have attracted considerable research interest. Wang \textit{et al.}~\cite{wang2018video} introduced FCN to this problem by taking adjacent pairs of frames as input. However, this method fails to learn sufficient spatiotemporal information with a limited number of input frames. To overcome this deficiency, Li \textit{et al.}~\cite{li2018flow} proposed to enhance the temporal coherence at the feature level by exploiting both motion information and sequential feature evolution encoding. Fan \textit{et al.}~\cite{fan2019shifting} proposed to captures video dynamics with a saliency-shift-aware module that learns human attention-shift. However, all the above methods rely on densely annotated video datasets, and none of them have ever attempted to reduce the dependence on dense labeling.

To the best of our knowledge, we are the first to explore the video salient object detection task by reducing the dependence on dense labeling. Moreover, we verify that the generated pseudo-labels can overcome the ambiguity in the labeling process to some extent, thus facilitating our model to achieve better performance.

\subsection{Video Object Segmentation}
Video object segmentation tasks can be divided into two categories, including semi-supervised video object segmentation~\cite{jain2014supervoxel, chockalingam2009adaptive} and unsupervised video object segmentation~\cite{tokmakov2017learningvideo, jain2017fusionseg}. Semi-supervised video object segmentation aims at tracking a target mask given from the first annotated frame in the subsequent frames, while unsupervised video object segmentation aims at detecting the primary objects through the whole video sequence automatically.
It should be noted that the supervised or semi-supervised video segmentation methods mentioned here are all for the test phase, and the training process of both tasks is fully supervised. The semi-supervised video salient object detection considered in this paper is aimed at reducing the labeling dependence of training samples during the training process.
Here, unsupervised video object segmentation is the most related task to ours as both tasks require no annotations during the inference phase. It can be achieved by graph cut~\cite{papazoglou2013fast}, saliency detection~\cite{wang2015saliency}, motion analysis~\cite{li2018unsupervised}, or object proposal ranking~\cite{lee2011key}. Recently, unsupervised video object segmentation methods have been mainly based on deep learning networks, such as two-stream architecture~\cite{jain2017fusionseg}, FCN network~\cite{cheng2017segflow}, and recurrent networks~\cite{tokmakov2017learningvideo}. However, most of the deep learning methods rely on a large quantity of pixel-wise labels for fully supervised training.

In this paper, we address the semi-supervised video salient object detection task using pseudo-labels with a few annotated frames. Although our proposed model is trained with semi-supervision, it is still well applicable to unsupervised video object segmentation.

\section{Our Approach}
\label{sec:approach}
In this section, we elaborate on the details of the proposed framework for semi-supervised video salient object detection, which consists of three major components. First, a residual connected refinement network is proposed to provide a spatial feature extractor and a pixel-wise classifier for salient object detection, which are respectively used for extracting spatial saliency features from raw input images and encoding the features to pixel-wise saliency maps with low-level cues connected to high-level features. Second, a non-locally enhanced recurrent module is designed to enhance the spatiotemporal coherence of the feature representation.
Finally, a flow-guided pseudo-label generation~(FGPLG) model, comprised of a modified RCRNet and an off-the-shelf FlowNet 2.0 model~\cite{ilg2017flownet}, is applied to generate in-between pseudo-labels from sparsely annotated video frames. With appropriate numbers of pseudo-labels, RCRNet with the NER module can be trained to capture the spatiotemporal information and generate accurate saliency maps for dense input frames.

\begin{figure}[t]
    \centerline{
       \includegraphics[width=0.48\textwidth]{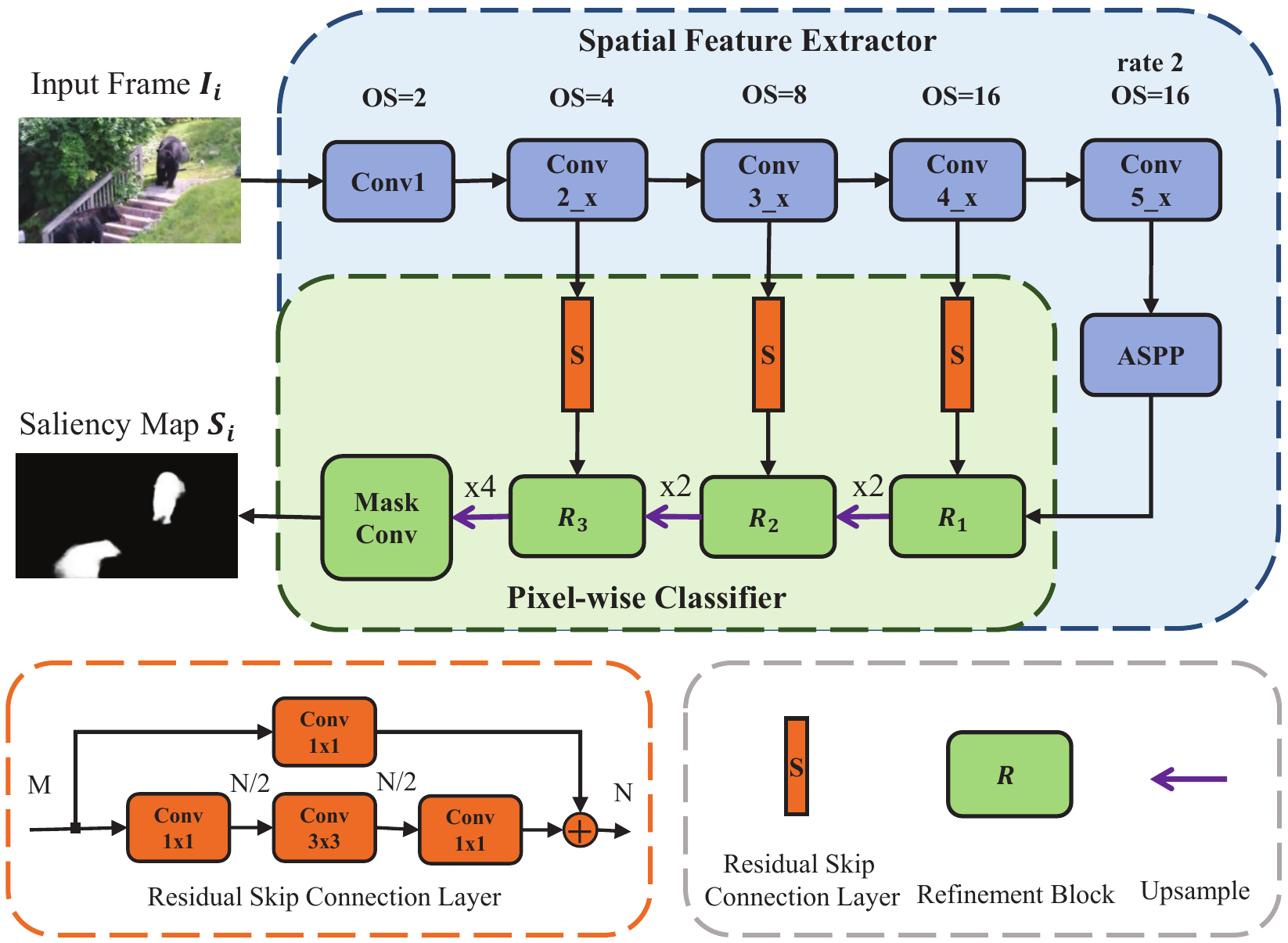}
    }
    \caption{The architecture of our refinement network with residual connection~(RCRNet). Here, `$\oplus$' denotes element-wise addition. Output stride~(OS) explains the ratio of the input image size to the output feature map size.
        }
    \label{fig:RCRNet}
\end{figure}

\begin{figure*}[ht]
    \begin{center}
       \includegraphics[width=0.91\textwidth]{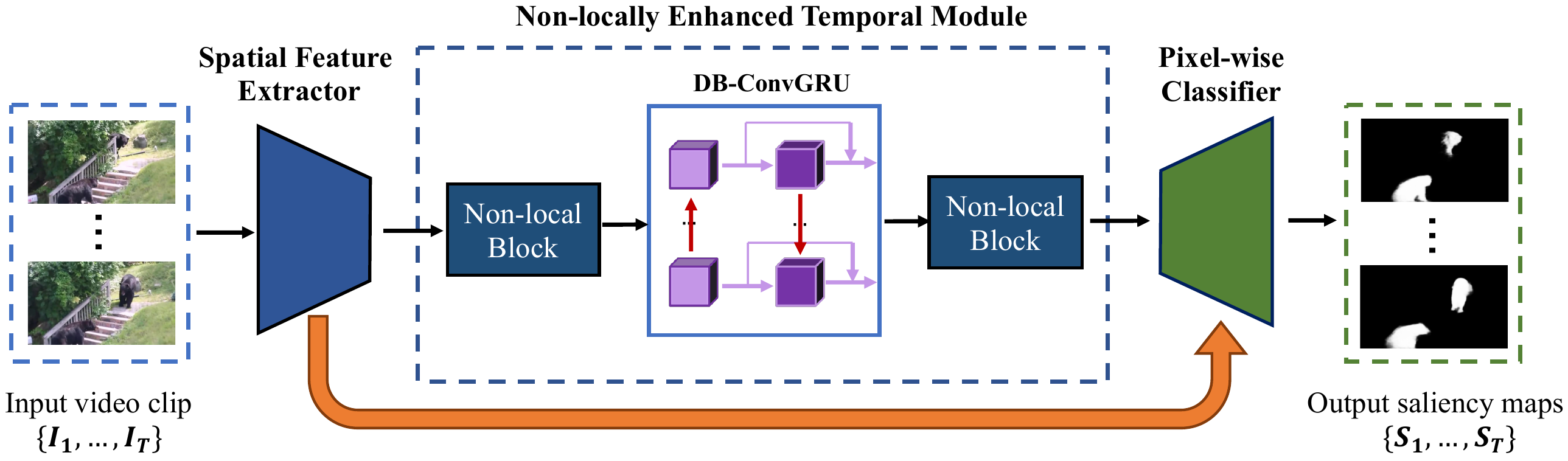}
    \end{center}
    \vspace{-3mm}
       \caption{The architecture of our proposed video salient object detection network~(RCRNet+NER). We incorporate a non-locally enhanced temporal module with our proposed RCRNet for spatiotemporal coherence modeling.}
    \label{fig:video_model}
\end{figure*}

\subsection{Refinement Network with Residual Connection}
Typical deep convolutional neural networks can extract high-level features from low-level cues of images, such as colors and textures, using a stack of convolutional layers and downsampling operations.
The downsampling operation obtains an abstract feature representation by gradually increasing the receptive field of the convolutional layers.
However, many spatial details are lost in this process. Without sufficient spatial details, pixel-wise prediction tasks, such as salient object detection, cannot precisely predict on object boundaries or small objects. Inspired by~\cite{li2017instance}, we adopt a refinement architecture to incorporate low-level spatial information in the decoding process for pixel-level saliency inference. As shown in Fig.~\ref{fig:RCRNet}, the proposed RCRNet consists of a spatial feature extractor $\mathcal{N}_{feat}$ and a pixel-wise classifier $\mathcal{N}_{seg}$ connected by three connection layers in different stages. The output saliency map $S$ of a given frame $I$ can be computed as
\begin{equation}
    S = \mathcal{N}_{seg}(\mathcal{N}_{feat}(I)).
\end{equation}

\noindent\textbf{Spatial Feature Extractor:} The spatial feature extractor is based on a ResNet-50~\cite{he2016deep} model. 
Specifically, we use the first five groups of layers of ResNet-50 and remove the downsampling operations in conv5\_x to reduce the loss of spatial information.
To maintain the same receptive field, we use dilated convolutions~\cite{yu2015multi} with $rate=2$ to replace the convolutional layers in the last layer. Then we attach an atrous spatial pyramid pooling~(ASPP)~\cite{chen2017rethinking} module to the last layer, which captures both the image-level global context and the multiscale spatial context. Finally, the spatial feature extractor produces features with $256$ channels and $1/16$ of the original input resolution~($\rm{OS}=16$).

\noindent\textbf{Pixel-wise Classifier:} The pixel-wise classifier is composed of three cascaded refinement blocks, each of which is connected to a layer in the spatial feature extractor via a connection layer. It is designed to mitigate the impact of the loss of spatial details during the downsampling process. Each refinement block takes as input the previous bottom-up output feature map and its corresponding feature map connected from the top-down stream. The resolution of these two feature maps should be consistent, so the upsampling operation is performed via bilinear interpolation when necessary. The refinement block works by first concatenating the feature maps and then feeding them to another $3\times3$ convolutional layer with $128$ channels. Motivated by~\cite{he2016deep}, a residual bottleneck architecture, named residual skip connection layer, is employed as the connection layer to connect low-level features to high-level ones. It downsamples the low-level feature maps from $M$ channels to $N=96$ channels and brings more spatial information to the refinement block. Residual learning allows us to connect the pixel-wise classifier to the pretrained spatial feature extractor without breaking its initial state~(e.g., if the weight of the residual bottleneck is initialized as zero).

\subsection{Non-locally Enhanced Recurrent Module}
\label{sec:nonlocal} Given a sequence of video clip $I_i, i=1,2,...,T$, video salient object detection aims at producing the saliency maps of all frames $S_i, i=1,2,...,T$. Although the proposed RCRNet specializes in spatial saliency learning, it still lacks spatiotemporal modeling for video frames. Thus, we further propose a non-locally enhanced temporal (NER) module, which consists of two non-local blocks~\cite{wang2018non} and a convolutional GRU~(ConvGRU)~\cite{ballas2015delving} module, to improve spatiotemporal coherence in high-level features. As shown in Fig.~\ref{fig:video_model}, incorporated with the NER module, RCRNet can be extended to video-based salient object detection.

Specifically, we first combine the features extracted from input video frames $\{I_i\}_{i=1}^T$ as $X = [X_1, X_2, ..., X_T].$ Here, [,.,] denotes the concatenation operation and the spatial feature $X_i$ of each frame $I_i$ is computed as $X_i=\mathcal{N}_{feat}(I_i)$. Then, the combined feature $X$ is fed into a non-local block. The non-local block computes the response at a position as a weighted sum of features at all positions for input feature maps. It can construct the spatiotemporal connection between the features of input video frames.


On the other hand, as a video sequence is composed of a series of scenes that are captured in chronological order, it is also necessary to characterize the sequential evolution of appearance contrast in the temporal domain.
Based on this, we propose to exploit ConvGRU~\cite{ballas2015delving} modules for sequential feature evolution modeling.
ConvGRU is an extension of traditional fully connected GRU~\cite{cho2014learning} that has convolutional structures in both input-to-state and state-to-state connections.
Let $\mathcal{X}_1, \mathcal{X}_2,...,\mathcal{X}_t$ denote the input to ConvGRU and $\mathcal{H}_1, \mathcal{H}_2, ...,\mathcal{H}_t$ stand for its hidden states. A ConvGRU module consists of a reset gate $\mathcal{R}_t$ and an update gate $\mathcal{Z}_t$. With these two gates, ConvGRU can achieve selective memorization and forgetting. Given the above definition, the overall updating process of ConvGRU unrolled by time can be listed as follows:
\begin{equation}
\begin{aligned}
    & \mathcal{Z}_t = \sigma(\mathcal{W}_{x z} * \mathcal{X}_t + \mathcal{W}_{h z} * \mathcal{H}_{t-1}), \\
    & \mathcal{R}_t = \sigma(\mathcal{W}_{x r} * \mathcal{X}_t + \mathcal{W}_{h r} * \mathcal{H}_{t-1}), \\
    & \mathcal{H}_t^{\prime} = \tanh(\mathcal{W}_{x h} * \mathcal{X}_t + \mathcal{R}_t \circ (\mathcal{W}_{h h} * \mathcal{H}_{t-1})), \\
    & \mathcal{H}_t = (1 - \mathcal{Z}_t) \circ \mathcal{H}_t^{\prime} + \mathcal{Z}_t \circ \mathcal{H}_{t-1},
\end{aligned}
\end{equation}
where `$*$' denotes the convolution operator and `$\circ$' denotes the Hadamard product. $\sigma(.)$ represents the sigmoid function and $\mathcal{W}$ represents the learnable weight matrices. For notational simplicity, the bias terms are omitted.

Motivated by~\cite{song2018pyramid}, we stack two ConvGRU modules with forward and backward directions to strengthen the spatiotemporal information exchanges between two directions. In this way, deeper bidirectional ConvGRU~(DB-ConvGRU) can memorize not only past sequences but also future ones. It can be formulated as follows:
\begin{equation}
\begin{aligned}
    \mathcal{H}_t^f &= {\rm ConvGRU}(\mathcal{H}_{t-1}^f, X_t), \\
    \mathcal{H}_t^b &= {\rm ConvGRU}(\mathcal{H}_{t+1}^b, \mathcal{H}_t^f), \\
    \mathcal{H}_t   &= \tanh(\mathcal{W}_{h f} * \mathcal{H}_t^f + \mathcal{W}_{h b} * \mathcal{H}_t^b),
\end{aligned}
\end{equation}
where $\mathcal{H}_t^f$ and $\mathcal{H}_t^b$ represent the hidden state from forward and backward ConvGRU units, respectively. $\mathcal{H}_t$ represents the final output of DB-ConvGRU. $X_t$ is the $t^{th}$ output feature from the non-local block.

As proven in~\cite{wang2018non}, more non-local blocks in general lead to better results. Thus, we attach another non-local block to DB-ConvGRU to further enhance spatiotemporal coherence.

\begin{figure*}[ht]
    \begin{center}
       \includegraphics[width=0.95\textwidth]{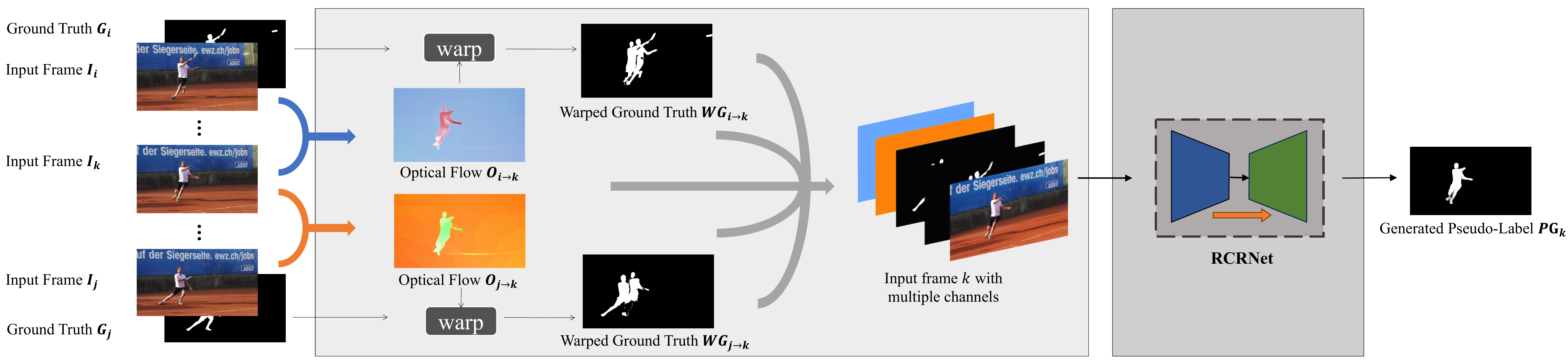}
    \end{center}
    \vspace{-3mm}
       \caption{The architecture of our proposed flow-guided pseudo-label generation model (FGPLG).}
    \label{fig:generation_model}
    \vspace{-1mm}
\end{figure*}

\subsection{Flow-Guided Pseudo-Label Generation Model}
\label{sec:generation}
Although the proposed RCRNet+NER has a great potential to produce saliency maps with spatiotemporal coherence. With only a few sparsely annotated frames, it can barely learn enough temporal information, which greatly reduces the temporal coherence of the resulting saliency maps. To solve this problem, we attempt to generate denser pseudo-labels from a few sparse annotations and train our video saliency model with both types of labels.

Given triplets of input video frames $\{I_i, I_k, I_j\}(i<k<j)$, the proposed FGPLG model aims at generating a pseudo-label for frame $I_k$ with ground truth $G_i$ and $G_j $ propagated from frame $I_i$ and $I_j$, respectively. First, it computes the optical flow $O_{i\rightarrow k}$ from frame $I_i$ to frame $I_k$ with the off-the-shelf FlowNet 2.0. The optical flow $O_{j\rightarrow k}$ is obtained in the same way. Then, the label of frame $I_k$ is estimated by applying a warping function to adjacent ground truth $G_i$ and $G_j$. Nevertheless, as we can see in Fig.~\ref{fig:generation_model}, the warped ground truth $WG_{i\rightarrow k}$ and $WG_{j\rightarrow k}$, are still too noisy to be used as supervisory information for practical training. Although the magnitude of optical flow $\left\|O_{i\rightarrow k}\right\|$ and $\left\|O_{j\rightarrow k}\right\|$ provide reasonable estimations of the motion mask of frame $i_k$, they cannot be employed as the estimated ground truth directly since not all the motion masks are salient. To further refine the estimated pseudo-label of frame $I_k$, another RCRNet is modified to accept a frame $I_k^+$ with 7 channels including RGB channels of frame $I_k$, adjacent warped ground truth $WG_{i\rightarrow k}$ and $WG{j\rightarrow k}$ and optical flow magnitude $\left\|O_{i\rightarrow k}\right\|$ and $\left\|O_{j\rightarrow k}\right\|$. With the above settings, a more reasonable and precise pseudo-label $PG_k$ of frame $I_k$ can be generated as:
\begin{equation}
    PG_k  = N_{seg}(N_{feat}(I_k^+)).
\end{equation}
Here, the magnitude of optical flow is calculated by first normalizing the optical flow into interval $[-1,1]$ and then computing its Euclidean norm.

The generation model can be trained with sparsely annotated frames to generate denser pseudo-labels.
In our experiments, we use a fixed interval $l$ to select sparse annotations for training. We take an annotation every $l$ frames, i.e., the interval between the $j^{th}$ and $k^{th}$ frame, and the interval between the $i^{th}$ and $k^{th}$ frame are both equal to $l$.
Experimental results show that the generation model designed in this way has a strong generalization ability. It can use the model trained by the triples sampled at larger interframe intervals to generate dense pseudo-labels of very high quality.

\begin{table*}
    \centering
    \resizebox{0.99\textwidth}{!}{%
    \begin{threeparttable}
    \begin{tabular}{c|r|ccc|ccccccc|ccc|ccc|c}
    \hline
    \multirow{2}{*}{Datasets} & \multirow{2}{*}{Metric} & \multicolumn{3}{c|}{I+C} & \multicolumn{7}{c|}{I+D} & \multicolumn{3}{c|}{V+U} & \multicolumn{4}{c}{V+D} \cr
    \cline{3-19}
     & & MC & RBD & MB+ & RFCN & DCL & DHS & DSS & MSR & DGRL & PiCA & SAG & GF & SSA & FCNS & FGRN & PDB & Ours$^*$ \cr
    \hline
    \multirow{2}{*}{VOS}   & $F_\beta^{max}\uparrow$ & 0.558 & 0.589 & 0.577 & 0.680 & 0.704 & 0.715 & 0.703 & 0.719 & 0.723 & \color[HTML]{32CB00}\textbf{0.734} & 0.541 & 0.529 & 0.669 & 0.681 & 0.714 & \color[HTML]{3166FF}\textbf{0.741} & \color[HTML]{FE0000}\textbf{0.856} \cr
                           & $S\uparrow$             & 0.612 & 0.652 & 0.638 & 0.721 & 0.728 & 0.783 & 0.760 & 0.764 & 0.776 & \color[HTML]{32CB00}\textbf{0.796} & 0.597 & 0.560 & 0.710 & 0.727 & 0.734 & \color[HTML]{3166FF}\textbf{0.797} & \color[HTML]{FE0000}\textbf{0.872} \cr
    \hline
    \multirow{2}{*}{DAVIS} & $F_\beta^{max}\uparrow$ & 0.488 & 0.481 & 0.520 & 0.732 & 0.760 & 0.785 & 0.775 & 0.775 & 0.758 & \color[HTML]{32CB00}\textbf{0.809} & 0.519 & 0.619 & 0.697 & 0.764 & 0.797 & \color[HTML]{3166FF}\textbf{0.849} & \color[HTML]{FE0000}\textbf{0.859} \cr
                           & $S\uparrow$             & 0.590 & 0.620 & 0.568 & 0.788 & 0.803 & 0.820 & 0.814 & 0.789 & 0.811 & \color[HTML]{32CB00}\textbf{0.844} & 0.663 & 0.686 & 0.738 & 0.757 & 0.838 & \color[HTML]{3166FF}\textbf{0.878} & \color[HTML]{FE0000}\textbf{0.884} \cr
    \hline
    \multirow{2}{*}{FBMS}  & $F_\beta^{max}\uparrow$ & 0.466 & 0.488 & 0.540 & 0.764 & 0.760 & 0.765 & 0.776 & 0.809 & 0.813 & \color[HTML]{32CB00}\textbf{0.823} & 0.545 & 0.609 & 0.597 & 0.752 & 0.801 & \color[HTML]{3166FF}\textbf{0.823} & \color[HTML]{FE0000}\textbf{0.861} \cr
                           & $S\uparrow$             & 0.567 & 0.591 & 0.586 & 0.765 & 0.772 & 0.793 & 0.793 & 0.835 & 0.832 & \color[HTML]{3166FF}\textbf{0.847} & 0.632 & 0.642 & 0.634 & 0.747 & 0.818 & \color[HTML]{32CB00}\textbf{0.839} & \color[HTML]{FE0000}\textbf{0.870} \cr

    \hline
    \end{tabular}
    \begin{tablenotes}
        \item[*] Note that our model is a semi-supervised learning model using only approximately $20\%$ ground truth labels for training.
    \end{tablenotes}
    \end{threeparttable}}
    \vspace{1mm}
    \caption{Comparison of quantitative results using maximum F-measure $F_{\beta}^{max}\uparrow$ (larger is better), S-measure $S\uparrow$ (larger is better). The best three results on each dataset are shown in \color[HTML]{FE0000}\textbf{red}, \color[HTML]{3166FF}\textbf{blue}\color{black}, and \color[HTML]{32CB00}\textbf{green}\color{black}, respectively. Symbols of model categories: I+C for image-based classic unsupervised or non-deep learning methods, I+D for image-based deep learning methods, V+U for video-based unsupervised methods, V+D for video-based deep learning methods. Refer to the supplemental document for more detailed results.}
    \label{tab:comp_quantity}
\end{table*}

\begin{figure*}[ht]
    \centerline{
        \includegraphics[width = 0.31\textwidth]{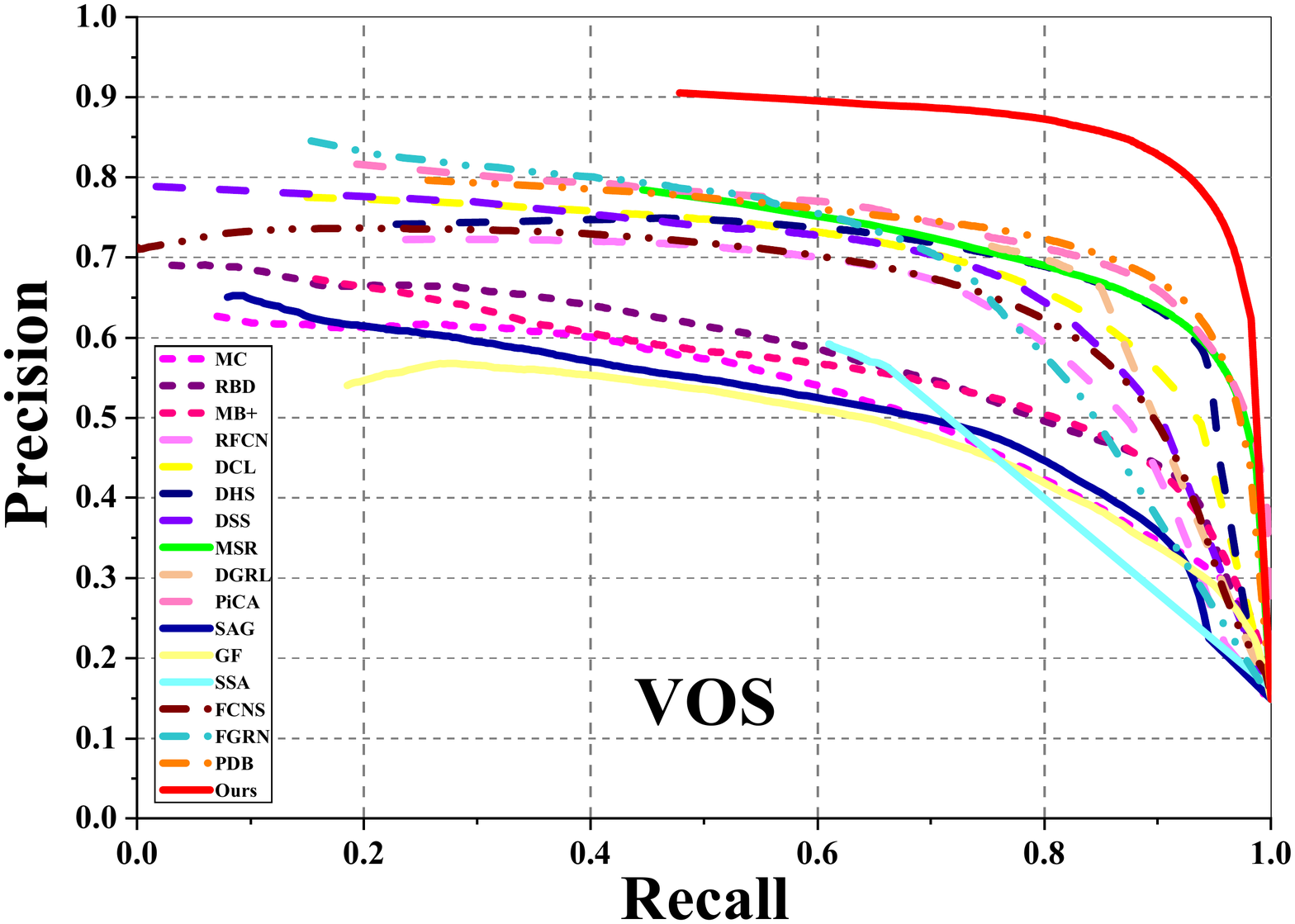}\hfill
        \includegraphics[width = 0.31\textwidth]{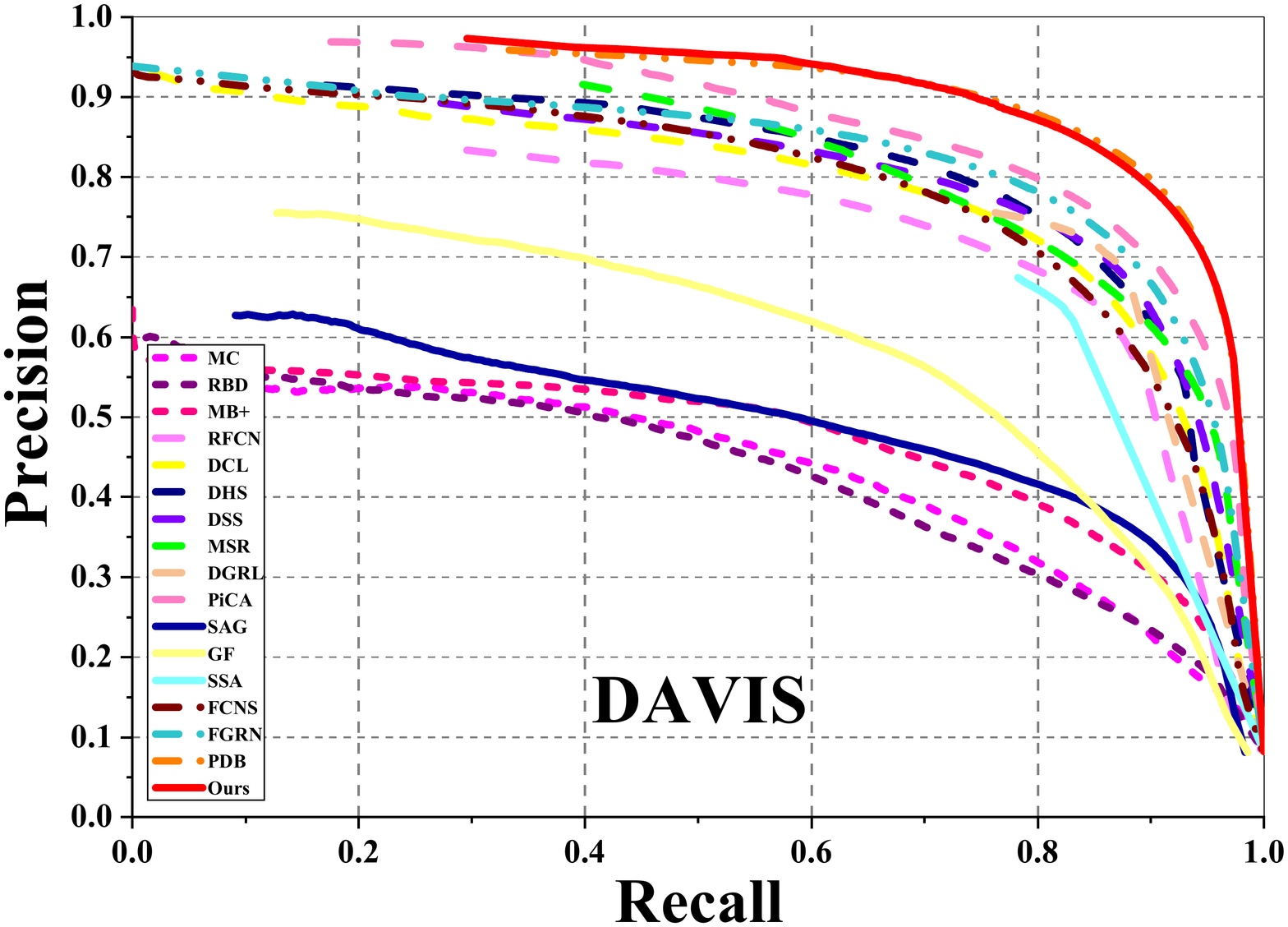}\hfill
        \includegraphics[width = 0.31\textwidth]{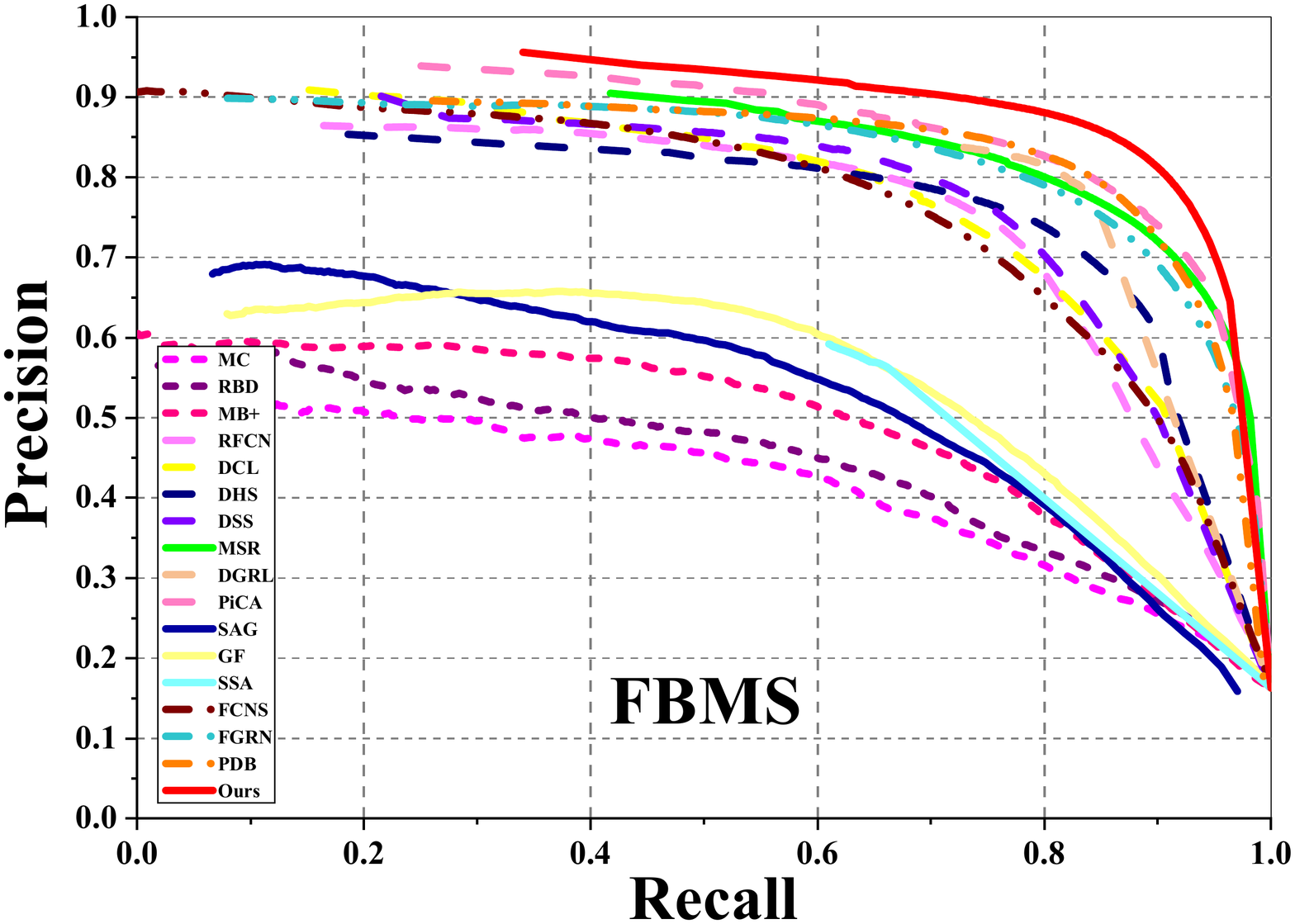}
    }
    \caption{Comparison of precision-recall curves of 15 saliency detection methods on the VOS, DAVIS and FBMS datasets. Our proposed RCRNet+NER consistently outperforms other methods across three testing datasets using only 20\% of ground truth labels.}
    \label{fig:cmp_pr_curve}
    \vspace{-2mm}
\end{figure*}
\section{Experimental Results}
\label{sec:experiment}

\subsection{Datasets and Evaluation}
We evaluate the performance of our method on three public datasets: VOS~\cite{li2018benchmark}, DAVIS~\cite{perazzi2016benchmark} and FBMS~\cite{brox2010object}. VOS is a large-scale dataset with 200 indoor/outdoor videos for video-based salient object detection. It contains 116,103 frames including 7,650 pixel-wise annotated keyframes. The DAVIS dataset contains 50 high-quality videos, with a total of 3,455 pixel-wise annotated frames. The FBMS dataset contains 59 videos, totaling 720 sparsely annotated frames. We evaluate our trained RFCN+NER on the test sets of VOS, DAVIS, and FBMS for the task of video salient object detection.

We adopt precision-recall curves~(PR), maximum F-measure and S-measure for evaluation. The F-measure is defined as
$F_{\beta} = \frac{(1+\beta^2)\cdot Precision \cdot Recall}{\beta^2 \cdot Precision + Recall}.$
Here, $\beta^2$ is set to 0.3 as done by most existing image-based models \cite{borji2015salient, li2017instance, hou2017deeply}. We report the maximum F-measure computed from all precision-recall pairs. The S-measure is a new measure proposed in~\cite{fan2017structure}, which can simultaneously evaluate both region-aware and object-aware structural similarity between a saliency map and its corresponding ground truth.

\subsection{Implementation Details}
\label{sec:implementation}
Our proposed method is implemented on PyTorch \cite{paszke2017automatic}, a flexible open source deep learning platform. First, we initialize the weights of the spatial feature extractor in RCRNet with an ImageNet~\cite{deng2009imagenet} pretrained ResNet-50~\cite{he2016deep}. Next, we pretrain the RCRNet using two image saliency datasets, i.e., MSRA-B~\cite{liu2011learning} and HKU-IS~\cite{li2015visual}, for spatial saliency learning.
For semi-supervised video salient object detection, we combine the training sets of VOS~\cite{li2018benchmark}, DAVIS \cite{perazzi2016benchmark}, and FBMS~\cite{brox2010object} as our training set. The RCRNet pretrained on image saliency datasets is used as the backbone of the pseudo-label generator. Then the FGPLG fine-tuned with a subset of the video training set is used to generate pseudo-labels. By utilizing the pseudo-labels together with the subset, we jointly train the RCRNet+NER, which takes a video clip of length $T$ as input, to generate saliency maps to all input frames.
Due to the limitation of machine memory, the default value of $T$ is set to 4 in our experiments.

During the training process, we adopt Adam~\cite{kingma2014adam} as the optimizer.
The learning rate is initially set to 1e-4 when training RCRNet, and is set to 1e-5 when fine-tuning RCRNet+NER and FGPLG. The input images or video frames are resized to $448\times448$ before being fed into the network in both training and inference phases. We use sigmoid cross-entropy loss as the loss function and compute the loss between each input image/frame and its corresponding label, even if it is a pseudo-label. In Section~\ref{seg:ablation_label_usage}, we explore the effect of different amount of ground truth~(GT) and pseudo-labels usage.
It shows that when we take one GT and generate one pseudo-label every five frames~(column `1 / 5' in Table~\ref{tab:ablation_label_usage}) as the new training set, RCRNet+NER can be trained to outperform the model trained with all ground truth labels on the VOS dataset. We use this setting when performing external comparisons with existing state-of-the-art methods.
In this setting, it takes approximately 10 hours to finish the whole training process on a workstation with an NVIDIA GTX 1080 GPU and a 2.4 GHz Intel CPU. In the inference phase, it takes approximately 37 ms to generate a saliency map for a $448\times448$ input frame, which reaches a real-time speed of 27 fps.

\subsection{Comparison with State-of-the-Art}
\label{sec:comparison}
We compare our video saliency model~(RCRNet+NER) against 16 state-of-the-art image/video saliency methods, including MC~\cite{jiang2013saliency}, RBD~\cite{zhu2014saliency}, MB+~\cite{zhang2015minimum}, RFCN~\cite{wang2016saliency}, DCL~\cite{li2016deep}, DHS~\cite{liu2016dhsnet}, DSS~\cite{hou2017deeply}, MSR~\cite{li2017instance}, DGRL~\cite{wang2018detect}, PiCA~\cite{liu2018picanet}, SAG~\cite{wang2015saliency}, GF~\cite{wang2015consistent}, SSA~\cite{li2018benchmark}, FCNS~\cite{wang2018video}, FGRN~\cite{li2018flow}, and PDB~\cite{song2018pyramid}. For a fair comparison, we use the implementations provided by the authors and fine-tune all the deep learning-based methods using the same training set, as mentioned in Section ~\ref{sec:implementation}.

\begin{figure*}[ht]
    \begin{center}
       \includegraphics[width=0.96\textwidth]{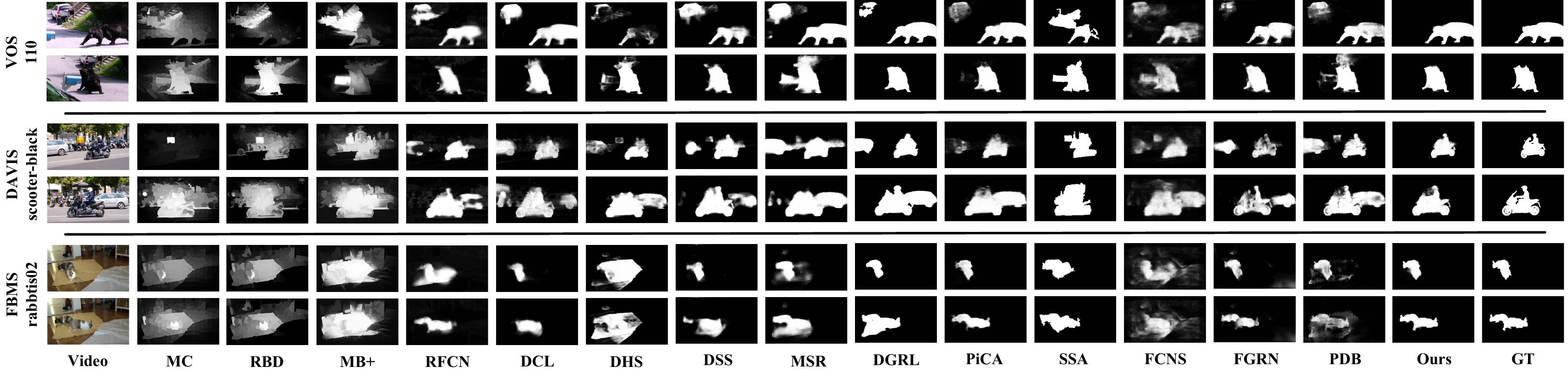}
    \end{center}
    \vspace{-3mm}
       \caption{Visual comparison of saliency maps generated by state-of-the-art methods, including our RCRNet+NER. The ground truth~(GT) is shown in the last column. Our model consistently produces saliency maps closest to the ground truth. Zoom in for details.}
    \label{fig:cmp_visual}
    \vspace{-3mm}
\end{figure*}

A visual comparison is given in Fig.~\ref{fig:cmp_visual}. As shown in the figure, RCRNet+NER can not only accurately detect salient objects but also generate precise and consistent saliency maps in various challenging cases. As a part of the quantitative evaluation, we show a comparison of PR curves in Fig.~\ref{fig:cmp_pr_curve}. Moreover, a quantitative comparison of maximum F-measure and S-measure is listed in Table~\ref{tab:comp_quantity}. As can be seen, our method can outperform all the state-of-the-art image-based and video-based saliency detection methods on VOS, DAVIS, and FBMS. Specifically, our RCRNet+NER improves the maximum F-measure achieved by the existing best-performing algorithms by 15.52\%, 1.18\%, and 4.62\% respectively on VOS, DAVIS, and FBMS, and improves the S-measure by 9.41\%, 0.68\%, 2.72\% accordingly. It is worth noting that our proposed method uses only approximately 20\% ground truth maps in the training process to outperform the best-performing fully supervised video-based method~(PDB), even though both models are based on the same backbone network~(ResNet-50).
\begin{figure}[t]
    \centerline{
        \includegraphics[width=0.4\textwidth]{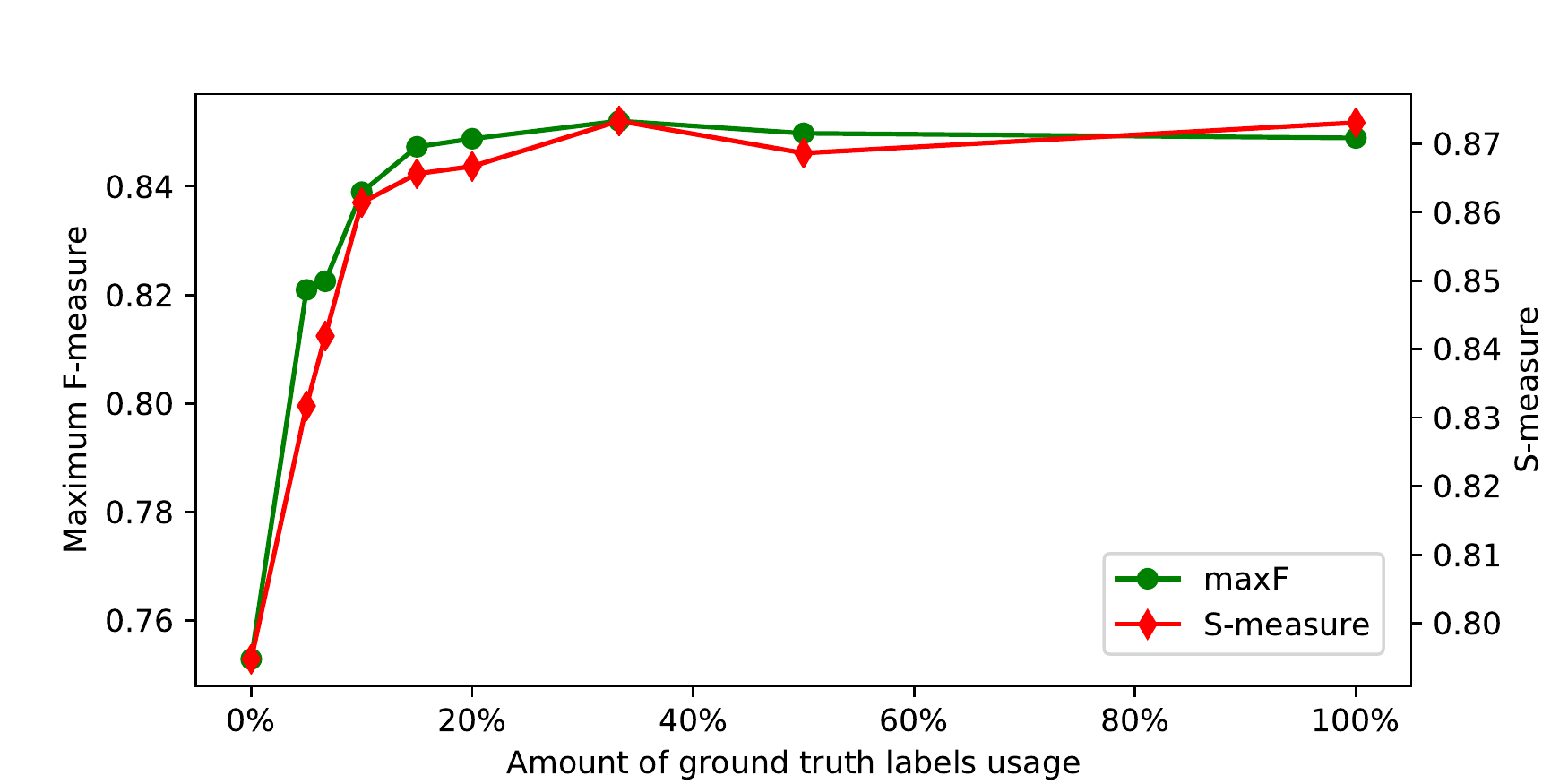}
    }
    \caption{Sensitivities analysis on the amount of ground truth labels usage.}
    \vspace{-1mm}
    \label{fig:ablation_label_usage}
\end{figure}

\begin{table}[h]
    \centering
    \resizebox{0.47\textwidth}{!}{%
    \begin{threeparttable}
    \begin{tabular}{c|c|c|c|c|c|c|c|c|c}
    \hline
    Labels & $m$ / $l$ & 0 / 1 & 0 / 2 & 0 / 5 & 1 / 5 & 4 / 5 & 0 / 20 & 7 / 20 & 19 / 20 \cr
    \hline
    \multirow{2}{*}{Proportion} & GT            & 100\% & 50\% & 20\% & 20\% & 20\% & 5\% & 5\%  & 5\%  \cr
    \cline{2-10}
                                & Pseudo        & 0\%   & 0\%  & 0\%  & 20\% & 80\% & 0\% & 35\% & 95\% \cr
    \hline
    \multirow{2}{*}{Metric} & $F_\beta^{max}\uparrow$ & 0.849 & 0.850 & 0.849 & \textbf{0.861} & 0.850 & 0.821 & 0.847 & 0.845 \cr
                            & $S\uparrow$       & 0.873 & 0.869 & 0.867 & \textbf{0.874} & 0.873 & 0.832 & 0.861 & 0.860 \cr
    \hline
    \end{tabular}
    \end{threeparttable}}
    \vspace{1mm}
    \caption{Some representative quantitative results on different amount of
    ground truth (GT) and pseudo-labels usage. Here, $l$ refers to the GT label interval, and $m$ denotes the number of pseudo-labels used in each interval. For example, `0 / 5' means using
    one GT every five frames with no pseudo-labels. `1 / 5' means using one GT and generating one pseudo-label every
    five frames. Refer to the supplemental document for more detailed analysis.}
    \label{tab:ablation_label_usage}
    \vspace{-3mm}
\end{table}

\subsection{Sensitivities to Different Amount of Ground Truth and Pseudo-Labels Usage}
\label{seg:ablation_label_usage}
As described in Section~\ref{sec:comparison}, RCRNet+NER achieves state-of-the-art performance using only a few GTs and generated pseudo-labels for training. To demonstrate the effectiveness of our proposed semi-supervised framework, we explore the sensitivities to different amount of GT and pseudo-labels usage on the VOS dataset. First, we take a subset of the training set of VOS by a fixed interval and then fine-tune the RCRNet+NER with it. By repeating the above experiment with different fixed intervals, we show the performance of RCRNet+NER trained with different number of GT labels in Fig.~\ref{fig:ablation_label_usage}. As shown in the figure, when the number of GT labels is severely insufficient~(e.g., 5\% of the origin training set), RCRNet+NER can benefit substantially from the increase in GT label usage.
An interesting phenomenon is that when the training set is large enough, the application of denser label data does not necessarily lead to better performance.
Considering that adjacent densely annotated frames share small differences, ambiguity is usually inevitable during the manual labeling procedure, which may lead to overfitting and affect the generalization performance of the model.

Then, we further use the proposed FGPLG to generate different number of pseudo-labels with different number of GT labels. Some representative quantitative results are shown in Table~\ref{tab:ablation_label_usage}, where we find that when there are insufficient GT labels, adding an appropriate number of generated pseudo-labels for training can effectively improve the performance. Furthermore, when we use 20\% of annotations and 20\% of pseudo-labels~(column `1 / 5' in the table) to train RCRNet+NER, it reaches the max $F_\beta=0.861$ and $S$-measure $=0.874$ on the test set of VOS, surpassing the one trained with all GT labels.
Even if trained with 5\% of annotations and 35\% of pseudo-labels~(column `7 / 20' in the table), our model can produce comparable results. This interesting phenomenon demonstrates that pseudo-labels can overcome labeling ambiguity to some extent. Moreover, it also indicates that it is not necessary to densely annotate all video frames manually considering redundancies.
Under the premise of the same labeling effort, selecting the sparse labeling strategy to cover more kinds of video content, and assisting with the generated pseudo-labels for training, will bring more performance gain.

\subsection{Ablation Studies}
\label{sec:ablation}
To investigate the effectiveness of the proposed modules, we conduct the ablation studies on the VOS dataset.

\noindent\textbf{The effectiveness of NER.}
As described in Section~\ref{sec:nonlocal}, our proposed NER module contains three cascaded modules, including a non-local block, a DB-ConvGRU module, and another non-local block. To validate the effectiveness and necessity of each submodule, we compare our RCRNet equipped with NER or its four variants on the test set of VOS. Here, we use one ground truth and one pseudo-label every five frames as the training set, to fix the impact of different amount of GT and pseudo-labels usage.
As shown in Table~\ref{tab:ablation_nonlocal}, $R_e$ refers to our proposed RCRNet with a non-locally enhanced module. By comparing the performance of the first three variants $R_a$, $R_b$, and $R_c$, we find that adding non-local blocks and DB-ConvGRU can create a certain level of performance improvement. On the basis of $R_c$, adding an extra non-local block~(i.e., $R_e$) can further increase 0.5\% w.r.t max F-measure. When compared with $R_d$ and $R_e$, we observe that DB-ConvGRU is indeed superior to ConvGRU as it involves deeper bidirectionally sequential modeling.

\vspace{1mm}
\noindent\textbf{The effectiveness of FGPLG.}
As mentioned in Section~\ref{sec:generation}, FGPLG model takes multiple channels as input to generate pseudo-labels, including image RGB channels, warped adjacent ground truth maps, and magnitude of optical flow.
To validate the effectiveness and necessity of each component, we train three separate RCRNet+NER with pseudo-labels generated by our proposed FGPLG including its two variants, each of which takes different channels as input.
Here, we use one ground truth and seven pseudo-labels every 20 frames as the training set for comparison.
It also includes the performance of model $G_a$, which is trained without pseudo-labels, as a baseline. As shown in Table~\ref{tab:ablation_generation}, the models trained with pseudo-labels~(i.e., $G_b$, $G_c$, and $G_d$) all surpass the baseline model $G_a$, which further validates the effectiveness of using pseudo-labels for training. On the basis of $G_b$, adding adjacent ground truth as input~(i.e., $G_c$) slightly improves the performance, while our proposed pseudo-label generator $G_d$ outperforms all the other variants with a significant margin by further exploiting adjacent ground truth through flow-guided motion estimation.

\begin{table}[h]
    \centering
    \resizebox{0.46\textwidth}{!}{%
    \begin{threeparttable}
    \begin{tabular}{c|c|c|c|c|c}
    \hline
    Methods                      & $R_a$        & $R_b$        & $R_c$        & $R_d$        & $R_e$        \cr
    \hline
    ConvGRU?                     &              &              &              & $\checkmark$ &              \cr
    DB-ConvGRU                   &              &              & $\checkmark$ &              & $\checkmark$ \cr
    first non-local block?       &              & $\checkmark$ & $\checkmark$ & $\checkmark$ & $\checkmark$ \cr
    second non-local block?      &              &              &              & $\checkmark$ & $\checkmark$ \cr
    \hline
    $F_\beta^{max}\uparrow$      & 0.846        & 0.853        & 0.856        & 0.857        &\textbf{0.861}\cr
    \hline
    $S\uparrow$       & 0.865        & 0.871        & 0.871        & 0.872        &\textbf{0.874}\cr
    \hline
    \end{tabular}
    \end{threeparttable}}
    \vspace{1mm}
    \caption{Effectiveness of non-locally enhanced recurrent module.}
    \vspace{-2.5mm}
    \label{tab:ablation_nonlocal}
\end{table}

\begin{table}[h]
    \centering
    \resizebox{0.46\textwidth}{!}{%
    \begin{threeparttable}
    \begin{tabular}{c|c|c|c|c}
    \hline
    Methods                      & $G_a$        & $G_b$        & $G_c$        & $G_d$        \cr
    \hline
    without label generation?    & $\checkmark$ &              &              &              \cr
    RGB channels?                &              & $\checkmark$ & $\checkmark$ & $\checkmark$ \cr
    adjacent ground truth?         &              &              & $\checkmark$ & $\checkmark$ \cr
    optical flow and GT warping? &              &              &              & $\checkmark$ \cr
    \hline
    $F_\beta^{max}\uparrow$      & 0.821        & 0.832        & 0.838        &\textbf{0.847}\cr
    \hline
    $S\uparrow$       & 0.832        & 0.854        & 0.860        &\textbf{0.861}\cr
    \hline
    \end{tabular}
    \end{threeparttable}}
    \vspace{1mm}
    \caption{Effectiveness of flow-guided label generation model.}
    \vspace{-2.5mm}
    \label{tab:ablation_generation}
\end{table}

\begin{table}[h]
    \centering
    \resizebox{0.47\textwidth}{!}{%
    \begin{threeparttable}
    \begin{tabular}{c|c|c|c|c|c|c|c|c}
    \hline
    \multirow{2}{*}{Dataset} & \multirow{2}{*}{Metric} & \multicolumn{7}{c}{Methods} \cr
    \cline{3-9}
    & & PDB & LVO & FSEG & LMP & SFL & FST & Ours$^*$ \cr
    \hline
    \multirow{2}{*}{DAVIS} & $\mathcal{J}\uparrow$ & 74.3 & 70.1 & 70.7 & 70.0 & 67.4 & 55.8 & \textbf{74.7} \cr
                           & $\mathcal{F}\uparrow$ & 72.8 & 72.1 & 65.3 & 65.9 & 66.7 & 51.1 & \textbf{73.3} \cr
    \hline
    FBMS                   & $\mathcal{J}\uparrow$ & 72.3 & 65.1 & 68.4 & 35.7 & 35.7 & 47.7 & \textbf{75.9} \cr
    \hline
    \end{tabular}
    \begin{tablenotes}
        \item[*] Note that our model is a semi-supervised learning model using only approximately $20\%$ ground truth labels for training.
    \end{tablenotes}
    \end{threeparttable}}
    \vspace{1mm}
    \caption{Performance comparison with 6 representative unsupervised video object segmentation methods on the DAVIS and FBMS datasets. The best scores are marked in \textbf{bold}.}
    \label{tab:comp_unsupervised}
    \vspace{-3mm}
\end{table}

\section{Performance on Unsupervised Video Object Segmentation}
Unsupervised video object segmentation aims at automatically separating primary objects from input video sequences. As described, its problem setting is quite similar to video salient object detection, except that it seeks to perform a binary classification instead of computing a saliency probability for each pixel. To demonstrate the advantages and generalization ability of our proposed semi-supervised model, we test the pretrained RCRNet+NER~(mentioned in Section~\ref{sec:experiment}) on the DAVIS and FBMS dataset without any pre-/post-processing and make a fair comparison with other 6 representative state-of-the-art unsupervised video segmentation methods, including FST~\cite{papazoglou2013fast}, SFL~\cite{cheng2017segflow}, LMP~\cite{tokmakov2017learningmotion}, FSEG~\cite{jain2017fusionseg}, LVO~\cite{tokmakov2017learningvideo} and PDB~\cite{song2018pyramid}. We adopt the mean Jaccard index $\mathcal{J}$~(intersection-over-union) and mean contour accuracy $\mathcal{F}$ as metrics for quantitative comparison on the DAVIS dataset according to its settings. For the FBMS dataset, we employ the mean Jaccard index $\mathcal{J}$, as done by previous works~\cite{song2018pyramid, li2018flow}. As shown in Table~\ref{tab:comp_unsupervised}, our proposed method outperforms the above methods on both the DAVIS and FBMS datasets, which implies that our method has a strong ability to capture spatiotemporal information from video frames and is applicable to unsupervised video segmentation.

\section{Conclusion}
In this paper, we propose an accurate and cost-effective framework for video salient object detection. Our proposed RCRNet equipped with a non-locally enhanced recurrent module can learn to effectively capture spatiotemporal information with only a few ground truths and an appropriate number of pseudo-labels generated by our proposed flow-guided pseudo-label generation model. We believe this will bring insights to future work on the manual annotation for video segmentation tasks. Experimental results demonstrate that our proposed method can achieve state-of-the-art performance on video salient object detection and is also applicable to unsupervised video segmentation. In future work, we will further explore the impact of the use of keyframe selection instead of interval sampling of GT labels on the performance of the proposed method.

\section*{Acknowledgements}
This work was supported by the State Key Development Program under Grant 2016YFB1001004, the National Natural Science Foundation of China under Grant No.U1811463, No.61702565 and No.61876045, the Department Science and Technology of Guangdong Province Fund under Grant No.2018B030338001, and was also sponsored by SenseTime Research Fund.

{\small
\bibliographystyle{ieee_fullname}
\bibliography{ICCV19_RCRNet_Saliency}
}
\includepdf[pages=1]{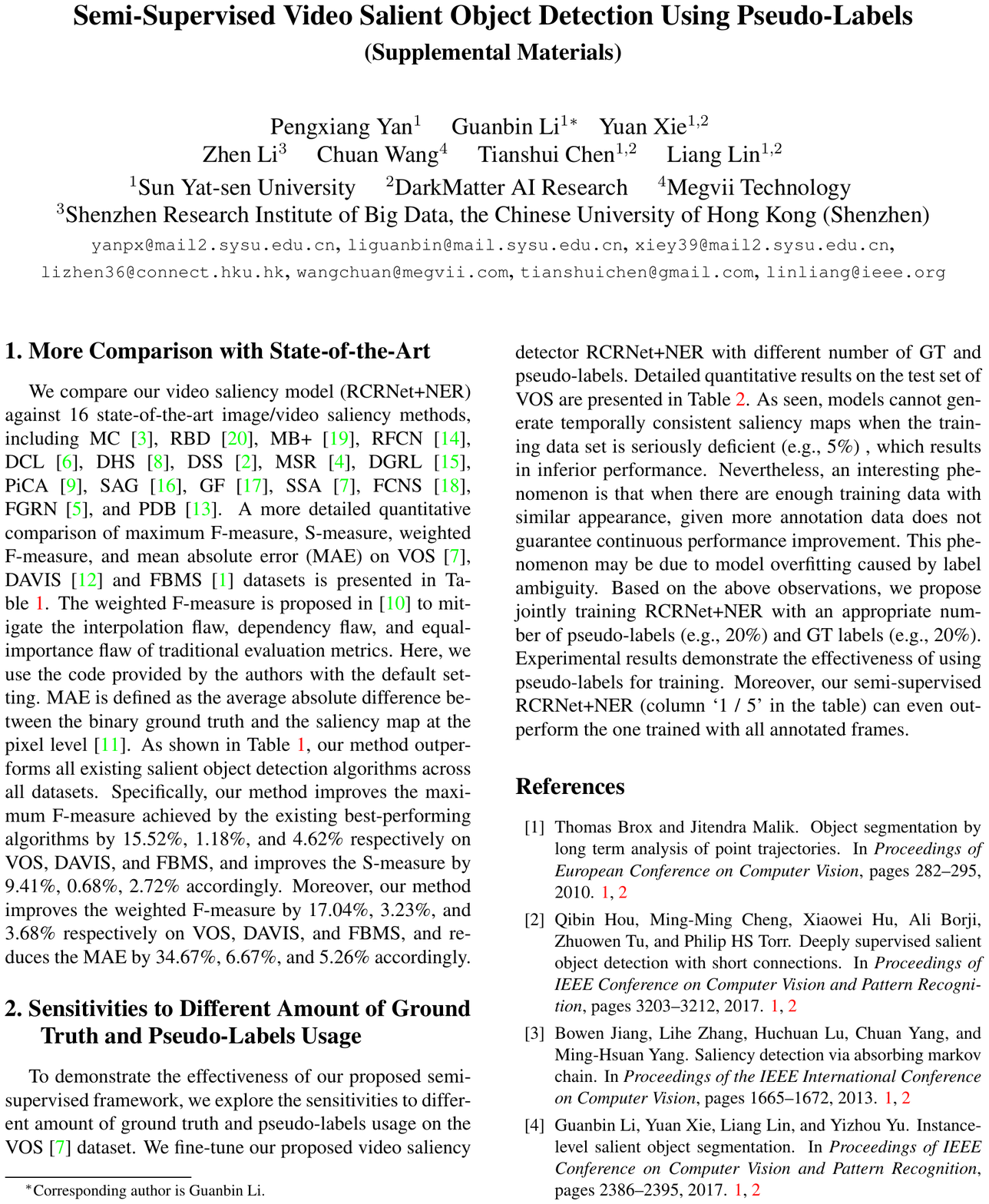}
\includepdf[pages=2]{ICCV19_RCRNet_Saliency-supp.pdf}
\includepdf[pages=3]{ICCV19_RCRNet_Saliency-supp.pdf}

\end{document}